\documentclass[sigconf]{acmart}

\usepackage{subcaption}
\usepackage{multirow}

\AtBeginDocument{%
  }

\acmSubmissionID{6120}

\begin{document}

\makeatletter
\def\@ACM@checkaffil{
    \if@ACM@instpresent\else
    \ClassWarningNoLine{\@classname}{No institution present for an affiliation}%
    \fi
    \if@ACM@citypresent\else
    \ClassWarningNoLine{\@classname}{No city present for an affiliation}%
    \fi
    \if@ACM@countrypresent\else
        \ClassWarningNoLine{\@classname}{No country present for an affiliation}%
    \fi
}
\makeatother

\title[PrefPaint: Enhancing Medical Image Inpainting through Expert Human Feedback]{PrefPaint: Enhancing Medical Image Inpainting through Expert Human Feedback} 



\author{Duy-Bao Bui}
\authornote{Both authors contributed equally to this research.}
\affiliation{%
  \institution{University of Science}
  \city{Ho Chi Minh}
  \country{Vietnam}
}
\affiliation{%
  \institution{Vietnam National University}
  \city{Ho Chi Minh}
  \country{Vietnam}
}
\email{}
\author{Hoang-Khang Nguyen}
\authornotemark[1]
\affiliation{%
  \institution{University of Science}
  \city{Ho Chi Minh}
  \country{Vietnam}
}
\affiliation{%
  \institution{Vietnam National University}
  \city{Ho Chi Minh}
  \country{Vietnam}
}
\email{}
\author{Thao Thi Phuong Dao}
\orcid{0000-0002-0109-1114}
\affiliation{%
  \institution{University of Science}
  \city{Ho Chi Minh}
  \country{Vietnam}
}
\affiliation{%
  \institution{Thong Nhat Hospital}
  \city{Ho Chi Minh}
  \country{Vietnam}
}
\email{}
\author{Kim Anh Phung}
\affiliation{%
  \institution{University of Cincinnati}
  \city{Cincinnati}
  \state{Ohio}
  \country{US}
}
\email{}
\author{Tam V. Nguyen}
\orcid{0000-0003-0236-7992}
\affiliation{%
  \institution{University of Dayton}
  \city{Dayton}
  \state{Ohio}
  \country{US}
}
\email{}
\author{Justin Zhan}
\affiliation{%
  \institution{University of Cincinnati}
  \city{Cincinnati}
  \state{Ohio}
  \country{US}
}
\email{}
\author{Minh-Triet Tran}
\orcid{0000-0003-3046-3041}
\affiliation{%
  \institution{University of Science}
  \city{Ho Chi Minh}
  \country{Vietnam}
}
\affiliation{%
  \institution{Vietnam National University}
  \city{Ho Chi Minh}
  \country{Vietnam}
}
\email{}
\author{Trung-Nghia Le}
\orcid{0000-0002-7363-2610}
\affiliation{%
  \institution{University of Science}
  \city{Ho Chi Minh}
  \country{Vietnam}
}
\affiliation{%
  \institution{Vietnam National University}
  \city{Ho Chi Minh}
  \country{Vietnam}
}
\email{}
\authornote{Corresponding author. Email: ltnghia@fit.hcmus.edu.vn}

\renewcommand{\shortauthors}{Anonymous authors}

\begin{abstract}

    Inpainting, the process of filling missing or corrupted image parts, has broad applications in medical imaging. However, generating anatomically accurate synthetic polyp images for clinical AI is a largely underexplored problem. In specialized fields like gastroenterology, inaccuracies in generated images can lead to false patterns and significant errors in downstream diagnosis. To ensure reliability, models require direct feedback from domain experts like oncologists. We propose PrefPaint, an interactive system that incorporates expert human feedback into Stable Diffusion Inpainting. By using D3PO instead of full RLHF, our approach bypasses the need for computationally expensive reward models, making it a highly practical choice for resource-constrained clinical settings. Furthermore, we introduce a streamlined web-based interface to facilitate this expert-in-the-loop training. Central to this platform is the Model Tree versioning interface, a novel HCI concept that visualizes the evolutionary progression of fine-tuned models. This interactive interface provides a smooth and intuitive user experience, making it easier to offer feedback and manage the fine-tuning process. User studies show that PrefPaint outperforms existing methods, reducing visual inconsistencies and generating highly realistic, anatomically accurate polyp images suitable for clinical AI applications.
    
\end{abstract}



\begin{teaserfigure}
  \includegraphics[width=\textwidth]{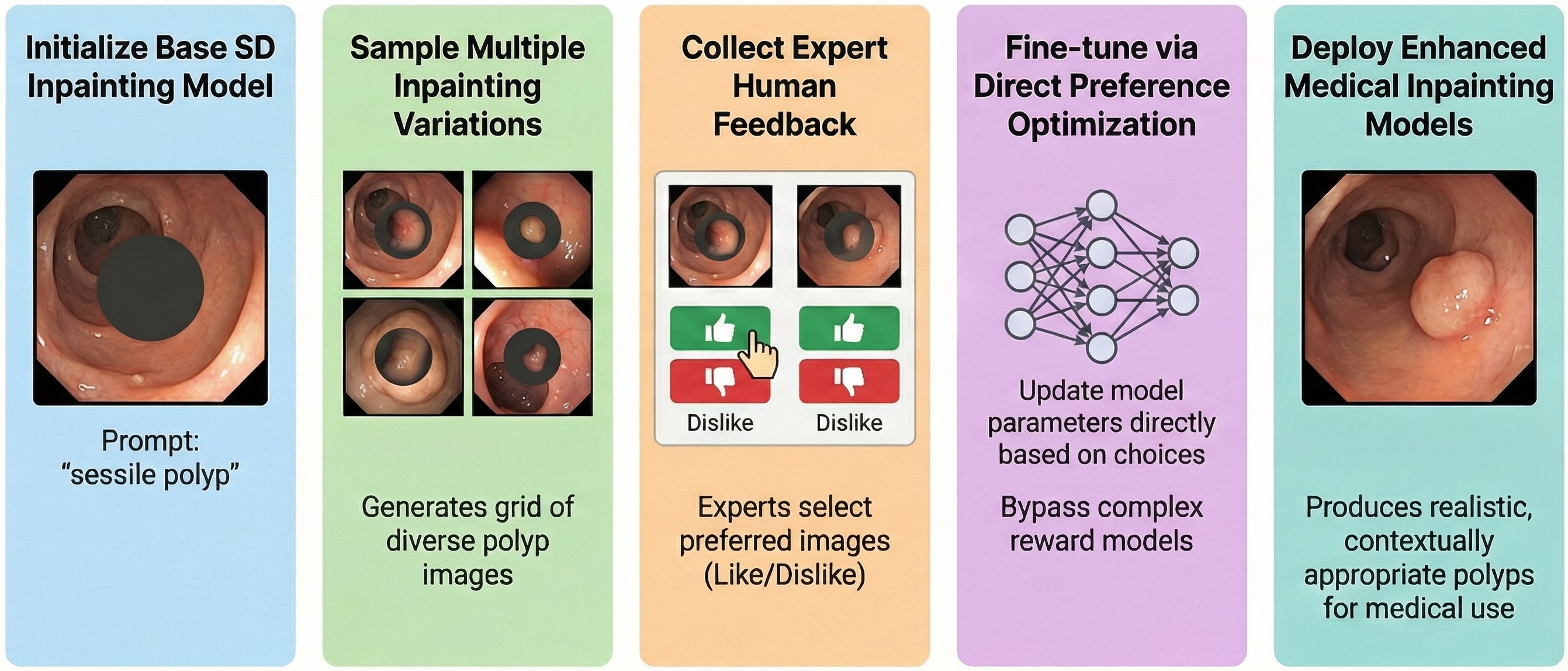}
  \caption{Proposed flow diagram illustrating the integration of human feedback in fine-tuning  medical image inpainting model.}
  \Description{}
  \label{fig:method-overview-medical-hf_flow}
\end{teaserfigure}

\maketitle

\section{Introduction}
Recent advancements in image generation models have achieved remarkable success in creating high-quality images from text descriptions \cite{karras2020analyzing,ramesh2021zero}. Various techniques, including Generative Adversarial Networks (GANs) \cite{goodfellow2014generative}, normalizing flows \cite{dinh2016density,rezende2015variational}, and diffusion models \cite{nichol2021glide,ramesh2022hierarchical}, have significantly advanced the capabilities of these systems. These models can now generate images that are both visually appealing and semantically accurate, attracting considerable interest for their potential applications and implications.

Inpainting, which involves filling in missing or corrupted parts of an image, and it downstream application, outpainting, which extends an image beyond its original borders, are essential image generation tasks. These techniques have numerous applications, including image restoration, content creation. However, inpainting medical images, especially those involving polyps, where precision and reliability are critical, presents distinct challenges compared to more general inpainting tasks. The problem of generating anatomically accurate synthetic polyp images for clinical AI is important, and underexplored. Polyps are precursors to colorectal cancer, and early detection is crucial for patient survival; yet, polyps are missed in up to 14\%-30\% of screening colonoscopies~\cite{jha2020kvasir}. Training robust clinical AI to detect these anomalies requires vast amounts of annotated data, but generating synthetic medical imagery without expert ground-truth validation often results in models learning incorrect anatomical patterns.

The lack of annotated polyp-images in training deep learning models requires to synthesize diverse images. However, training models based entirely on unannotated data can lead to generated data that may not match reality, for example, this tumor is not real, leading to down-stream training of the model wasting time, etc. To safely generate diverse, realistic training data, generative models must directly integrate feedback from medical experts. While Reinforcement Learning from Human Feedback (RLHF)~\cite{christiano2017deep} has proven effective for large language models~\cite{openai2023gpt}, it traditionally requires training an intermediate reward model, a process demanding extensive datasets, optimal architecture, and massive computational overhead, making the process both time-consuming and costly.


To bridge the gap between complex machine learning workflows and clinical end-users, we propose PrefPaint. Our system enables integrating human expert feedback into the training process of Stable Diffusion Inpainting (Fig.~\ref{fig:method-overview-medical-hf_flow}) to take advantage of knowledge from medical experts, thus generated polyp-images receive accurate feedback from doctors with tumor expertise. To address the issue of high computational overhead, we employ D3PO~\cite{yang2024using} for utilizing resource-constrained clinical settings because this approach allows for the direct updating of model parameters based on binary expert preferences, completely bypassing the expensive reward model. By directly incorporating insights from medical professionals, our approach can enhance the accuracy and reliability of image generation processes, which are critical in medical diagnostics and treatment planning.

We also design and develop a user-friendly and intuitive web-based interactive interface, even for medical professionals who may not have a strong IT background. This platform can facilitate the collection of expert feedback and effectively track the current progress and performance improvements of the model over time, ensuring that the system can be easily used by individuals with medical expertise. A cornerstone of our system is the Model Tree versioning interface, a novel HCI concept that provides clinicians with a clear, hierarchical visualization of model provenance and iterative improvements over time. By providing a seamless interface, our system can gather valuable input from users, which can further refine and improve the models' performance in real-world applications. 

Extensive experiments were conducted to assess our models for both quantitative and qualitative evaluations and user studies. Our proposed inpainting model significantly outperformed existing state-of-the-art methods, producing more realistic and contextually appropriate depictions of polyps.

Our main contributions are as follows:
\begin{itemize}
    \item We address the critical, underexplored problem of generating anatomically accurate synthetic polyp images for clinical AI by integrating expert human feedback, offering a highly practical and resource-efficient solution for clinical environments.

    \item We introduce a responsive, web-based interactive system tailored for medical professionals, which features a novel HCI concept of a Model Tree versioning interface to intuitively visualize the evolutionary progression and provenance of fine-tuned models.

    \item We rigorously evaluate our proposed method through comprehensive quantitative metrics and user studies, demonstrating that it significantly outperforms existing state-of-the-art methods by reducing visual inconsistencies and generating highly realistic, contextually appropriate medical imagery.
\end{itemize}

\section{Related Work}

\subsection{Image Inpainting}

Denoising diffusion probabilistic models \cite{sohl2015deep, ho2020denoising}, have emerged as powerful tools for generating a variety of data types. These models have been successfully applied in domains such as image synthesis \cite{ramesh2021zero,saharia2022photorealistic}, video generation \cite{ho2022imagen,li2023finedance}, and robotics control systems \cite{ajay2022conditional}. Particularly, text-to-image diffusion models \cite{ramesh2021zero,saharia2022photorealistic} have enabled the creation of highly realistic images from textual descriptions, opening new possibilities in digital art and design. Recent research has aimed at refining the guidance of diffusion models for more precise control over the generative process. In this work, we utilize Stable Diffusion Inpainting \cite{ramesh2022hierarchical} to generate images based on specific prompts and masks.

Inpainted missing regions within an image need to harmonize with the rest of the image and be semantically reasonable. Inpainting approaches \cite{yu2019free,liu2020rethinking,aot,lama} thus require strong generative capabilities and need to handle various forms of masks such as thin or thick brushes, squares, or even extreme masks where the vast majority of the image is missing. This is highly challenging since existing approaches train with a certain mask distribution, which can lead to poor generalization to novel mask types. In this work, we investigate an alternative generative approach for inpainting, aiming to design an approach that requires no mask-specific training.

\begin{figure*}[t!]
    \centering
    \includegraphics[width=\linewidth, trim={3mm 22mm 2mm 27mm}, clip]{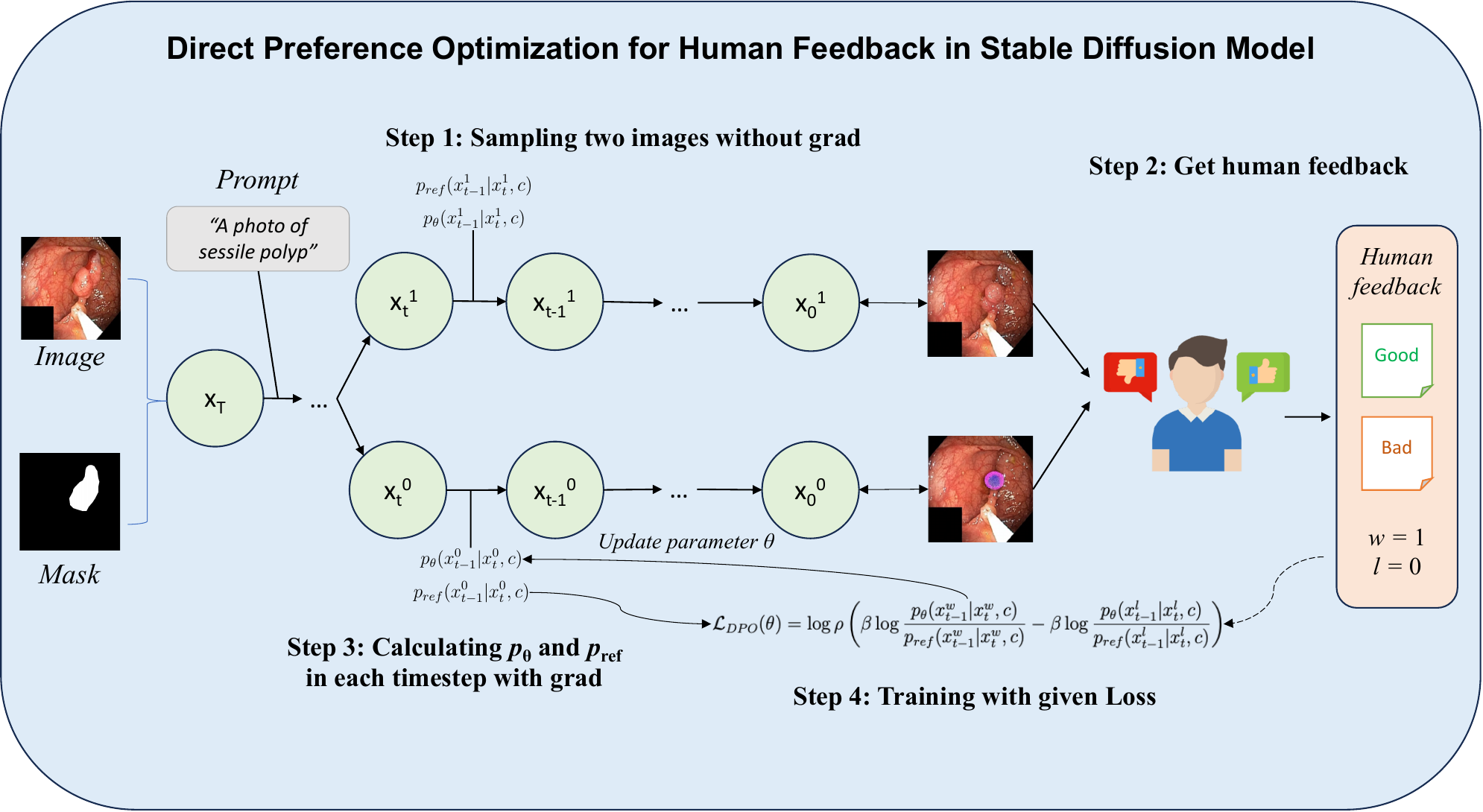}
    \caption{Overview of our system for medical image inpainting. Following D3PO~\cite{yang2024using}, the inpainting model generates two corresponding images based on the provided prompts. Guided by specific task requirements, such as improving prompt-image alignment or refining aesthetic quality, human evaluators select the preferred image. Leveraging this human feedback, our method directly updates the inpainting model’s parameters without necessitating the training of a reward model.}
    \label{fig:method-rl-weighted-overview}
\end{figure*}

\subsection{Human Feedback in Fine-tuning Models}
Reinforcement Learning from Human Feedback (RLHF) has emerged as a prominent strategy in machine learning, particularly for objectives that are complex or difficult to define explicitly. This technique has proven instrumental in various applications, from gaming \cite{bai2022training,christiano2017deep} to advanced robotics \cite{casper2023open,ziegler2019fine}. The integration of RLHF into the development of large language models (LLMs) represents a significant milestone in the field, with notable models such as OpenAI’s GPT-4 \cite{openai2023gpt}, Anthropic’s Claude \cite{anthropic2023introducing}, Google’s Bard \cite{aydin2023google}, and Meta’s Llama 2-Chat \cite{touvron2023llama} utilizing this approach to enhance performance and relevance. The effectiveness of RLHF in refining LLM behavior to align more closely with human values, such as helpfulness and harmlessness, has been extensively studied \cite{bai2022training,ziegler2019fine}. The technique has also proven beneficial in focused tasks like summarization, where models are trained to condense extensive information into concise representations \cite{stiennon2020learning}. Recent research has explored Reinforcement Learning from AI Feedback (RLAIF) \cite{lee2023rlaif} as an alternative to RLHF for model fine-tuning, offering convenience and efficiency by replacing human feedback with AI-generated feedback. However, for tasks such as assessing hand generation normality or image aesthetic appeal, reliable judgment models are currently lacking. In addition, conventional RLHF techniques often rely on computationally expensive reward models, making them challenging to apply to complex vision training tasks.

In reinforcement learning, there has been growing interest in exploring policies derived from preferences rather than explicit rewards. Contextual Dueling Bandit framework \cite{dudik2015contextual,yue2012k} introduces the concept of a von Neumann winner, shifting the focus away from directly pursuing an optimal policy based on rewards. Preference-based Reinforcement Learning \cite{busa2014preference} learns from binary preferences inferred from a cryptic scoring function instead of explicit rewards. Recently, Direct Preference Optimization (DPO) \cite{rafailov2024direct} has been proposed, which fine-tunes LLMs directly using preferences. This approach leverages the correlation between reward functions and optimal policies, effectively addressing the challenge of constrained reward maximization in a single phase of policy training. In this work, we employ a DPO-based training solution to reduce the complexity of training image inpainting models.

\section{Proposed System}

\subsection{Integration of Human Feedback in Training Inpainting Model}

To address the challenges in medical image inpainting, integrating human feedback into the training process is essential. Human feedback, particularly from medical professionals with expertise in tumor and cancer diagnosis, provides invaluable insights that help fine-tune the models. This feedback loop ensures that the generated images align more closely with real medical conditions, enhancing their accuracy and reliability. The pipeline of integrating the human feedback to fine-tune the Stable Diffusion Inpainting model, presented in Fig.~\ref{fig:method-overview-medical-hf_flow}, involves several key steps:

\textbf{Initial Model Training}: The inpainting model is initially trained on the available medical image datasets. Although these datasets may be limited and lack comprehensive ground truth annotations, they provide a starting point for the model.
    
\textbf{Generation of Inpainting Images}: The trained model generates inpainted images based on the input medical images with missing regions. These generated images are then reviewed by medical experts.

\textbf{Human Feedback Collection}: Medical experts provide feedback on the accuracy and realism of the inpainted images. This feedback includes identifying inaccuracies, suggesting improvements, and annotating areas where the model's predictions do not align with real medical observations.

\textbf{Model Fine-Tuning}: The inpainting model is fine-tuned using reinforcement learning techniques. This process iteratively improves the model's performance by optimizing it to generate images that receive higher scores from the reward model.

Crucially, we employ a training solution based on D3PO~\cite{yang2024using} rather than traditional RLHF, which is a practical and necessary choice for resource-constrained clinical settings. Typical medical institutions lack the massive compute clusters required to train complex reward models. By conceptualizing the denoising process as a multi-step Markov Decision Process (MDP), D3PO directly fine-tunes the inpainting model using binary human feedback, saving significant computational resources while strictly adhering to expert clinical guidance. We redefine states, transition probabilities, and policy functions as in Fig.~\ref{fig:method-rl-weighted-overview}. The model is trained using the DDPO loss:
\begin{equation}
\begin{aligned}
    \mathcal{L}(\theta) = -\mathbb{E}_{(x_t,\sigma_w,\sigma_l)}\Bigg[\log \rho\Bigg(&\beta\log\frac{p_\theta(x_{t-1}^w|x_t^w,c) }{p_\mathrm{ref}(x_{t-1}^w|x_t^w,c)} \\
    &- \beta\log\frac{p_\theta(x_{t-1}^l|x_t^l,c) }{p_\mathrm{ref}(x_{t-1}^l|x_t^l,c)} \Bigg) \Bigg],
\end{aligned}
\label{eq:method-policyoptimize-loss_d3po}
\end{equation}
where tuple $\boldsymbol{c}=\{p, i, m\}$ includes prompt $\boldsymbol{p}$, image $\boldsymbol{i}$, mask $\boldsymbol{m}$, sample $\boldsymbol{x}_{T}\sim \mathcal{N}(\mathbf{0},\mathbf{I})$, $t \in \left[0, T-1\right]$, and $\sigma_w$ denotes the segment preferred over another segment $\sigma_l$.

\subsection{Interactive System Architecture and Implementation}

To seamlessly integrate expert human feedback into the fine-tuning of diffusion models, we developed a highly responsive web-based interface. Because model training and inference are computationally intensive tasks, ensuring fluid, uninterrupted human-computer interaction requires a robust and scalable architecture. The system is hosted on a machine equipped with an NVIDIA RTX 3090 GPU, 32GB of RAM, and a 100GB SSD, providing the necessary computational bandwidth to prevent system lag during user studies and experiments.

\begin{figure}[t!]
    \centering
    \includegraphics[width=\linewidth]{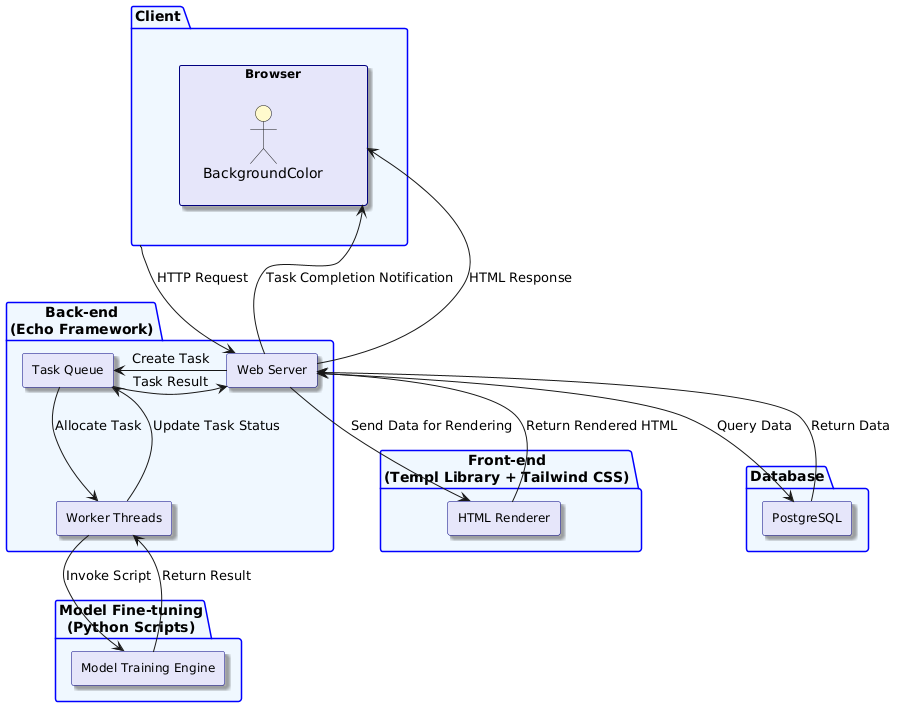}
    \caption{System architecture diagram.}
    \label{fig:webapp-diagram}
    \Description{A diagram.}
\end{figure}

Our platform employs a modern Server-Side Rendering (SSR) architecture built on Golang, divided into three cohesive components optimized for UI responsiveness (Fig.~\ref{fig:webapp-diagram}).

\subsubsection{Interactive Front-end and Robust Back-end}

We construct the user interface using templ, a strongly-typed, component-based templating engine that encapsulates HTML and interaction logic into reusable modules. By natively integrating with Golang, templ minimizes runtime errors and accelerates UI delivery. For styling, we employ Tailwind CSS, utilizing its just-in-time (JIT) compilation to generate lightweight, on-demand utility classes. This combination strictly optimizes UI payload sizes, ensuring rapid rendering and seamless visual updates for the user.

The core server relies on the lightweight echo routing framework to efficiently handle HTTP requests, mitigating common security pitfalls while significantly reducing boilerplate code. The back-end interfaces seamlessly with a PostgreSQL database, enabling high-speed Create-Read-Update-Delete (CRUD) operations that feed real-time data directly to the front-end renderer.

\subsubsection{Asynchronous Task Management for Model Fine-Tuning} 

A fundamental HCI challenge in interactive machine learning platforms is preventing the UI from freezing during heavy GPU operations. To guarantee uninterrupted interaction, we implemented a multi-threaded architecture driven by a First-In, First-Out (FIFO) task queue.

When a user initiates an action, such as sampling, fine-tuning, or inference, the request is instantly enqueued, and the user interface remains fully active. Dedicated worker threads, operating entirely independently of the main web thread, asynchronously pick up these tasks and spawn child processes to execute the required Python training scripts. The system continuously tracks and broadcasts task states (pending, processing, and finished) back to the user interface. This robust separation of concerns ensures the platform scales effectively while providing users with transparent, real-time feedback without ever blocking their workflow.

\subsection{Interactive System Workflow and UI}
\label{sec:UI}

To effectively integrate medical experts into the machine learning loop, our platform is designed to minimize cognitive load while providing robust control over model training. The interface is divided into functional workflows that seamlessly guide users through onboarding, iterative fine-tuning, inference, and system monitoring.

\subsubsection{Onboarding and User Guidance}

To accommodate medical professionals who may lack extensive technical backgrounds, the platform features a comprehensive and intuitive user guide. This module provides clear navigation instructions, ensuring users can rapidly familiarize themselves with the system's capabilities and confidently engage with the training workflow.

\subsubsection{Fine-Tune Station and Provenance Tracking}

\paragraph{Model Tree. } Iterative fine-tuning naturally generates multiple versions of a model. To help users track this evolutionary progression, we visualize model provenance using a top-down hierarchical tree. The root node represents the initial baseline model, with descending branches mapping subsequent fine-tuned iterations. This visual structure allows users to systematically track the developmental trajectory of their models, highlighting the cumulative impact of expert feedback over time.

\paragraph{Sampling and Feedback Collection. } Interacting with any node in the tree allows users to either run inference on real-world data or initiate further fine-tuning. During the feedback collection phase, the interface presents generated image samples alongside a strictly binary "like" or "dislike" rating system. By mapping preferences to simple values (0 for like, -1 for dislike), this streamlined approach intentionally minimizes distractions and cognitive overhead, enabling experts to provide robust, high-volume feedback efficiently before submitting it to the training engine.

\paragraph{Interactive Inference and Showcase. } The inference interface allows users to evaluate models practically by uploading medical images and interactively drawing masks over regions requiring modification. Users can adjust the brush size for precise targeting and input descriptive text prompts to guide the generative process. To support qualitative evaluation, a dedicated Showcase gallery persistently displays historical inputs, applied text prompts, and the resulting output images, allowing users to easily compare transformations side-by-side.

\paragraph{Task Manager and System Transparency. } Machine learning workflows inherently involve variable latency; for instance, inference requests may take approximately 10 seconds, while model fine-tuning can require up to 20 minutes. To proactively manage user expectations and maintain system transparency, we implemented a real-time Task Manager. This interface provides live tracking of all asynchronous requests, displaying the current processing stage of each task. This ensures users remain continuously informed of backend operations, preventing them from feeling left in the dark during long computational processes.

\subsubsection{User Guide Page}
\begin{figure}[t!]
        \centering
        \includegraphics[width=\linewidth]{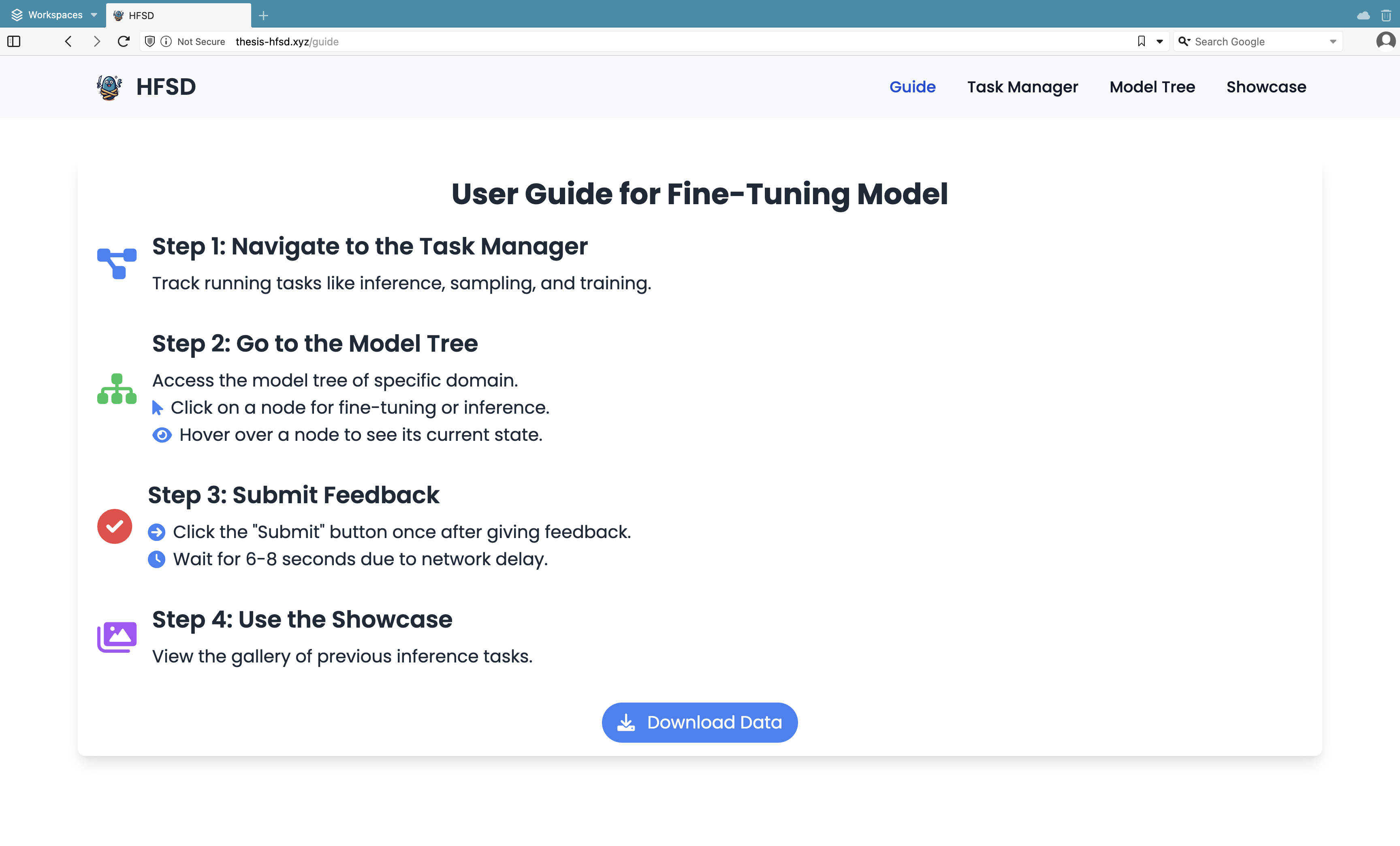}
    \vspace{-10mm}
    \caption{%
        Simple guide to web-based interface.
    }\label{fig:webapp-intro}
    \Description{A page landing page.}
    \vspace{-5mm}
\end{figure}

This page provides detailed instructions and information to help users navigate the website and utilize its features effectively (see Fig.~\ref{fig:webapp-intro}). The user guide is designed to be comprehensive and user-friendly, ensuring that users can quickly become familiar with the system and make the most of the resources and tools we offer. This intuitive navigation and clear guidance are crucial for enhancing the overall user experience and facilitating engagement with our research.

\subsubsection{Fine-tune Station}
\begin{figure}[t!]
    \begin{minipage}[t]{\linewidth}
        \centering
        \includegraphics[width=\linewidth]{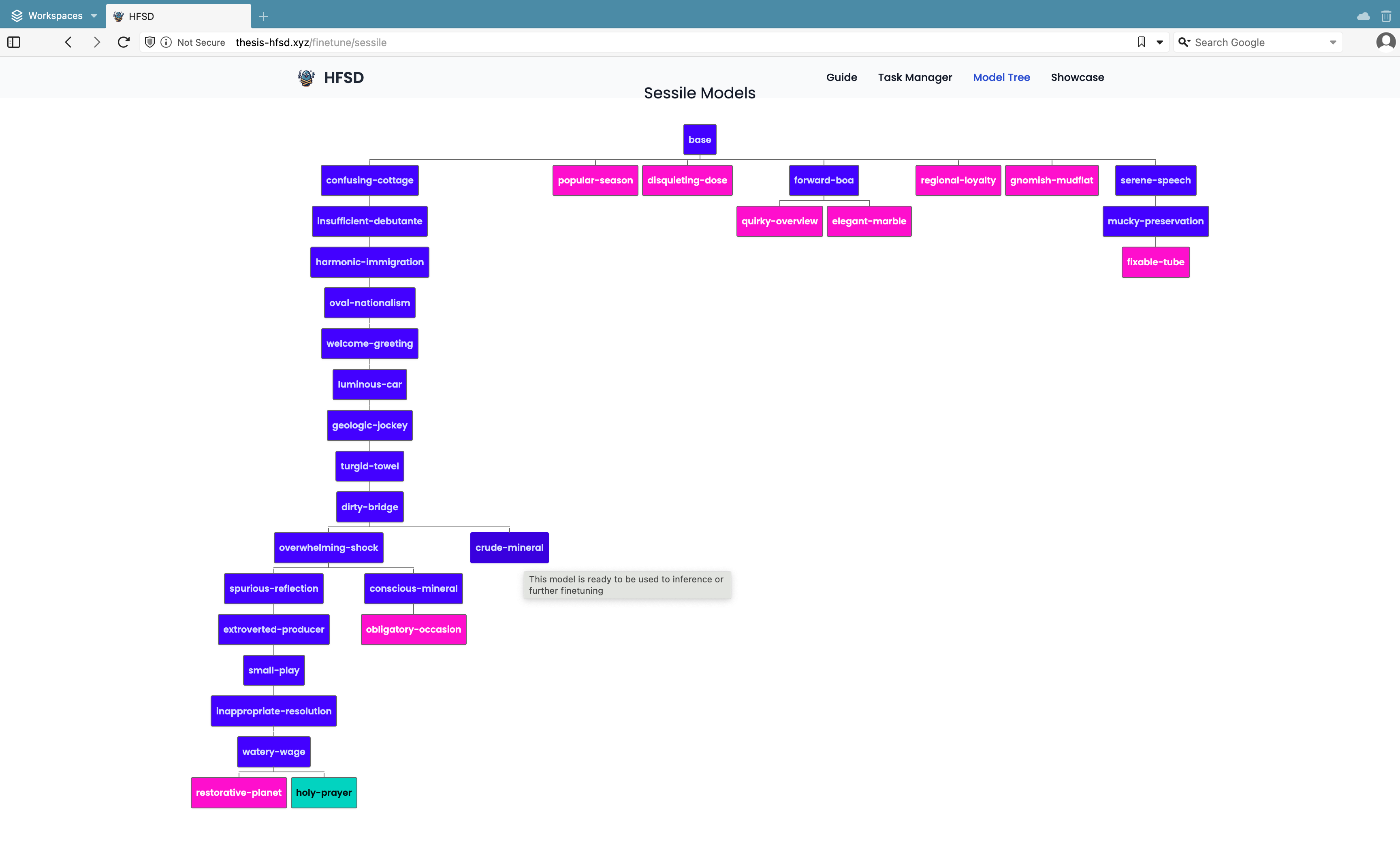}
    \end{minipage}
    \hspace{0.1cm}
    \begin{minipage}[t]{\linewidth}
        \centering
        \includegraphics[width=\linewidth]{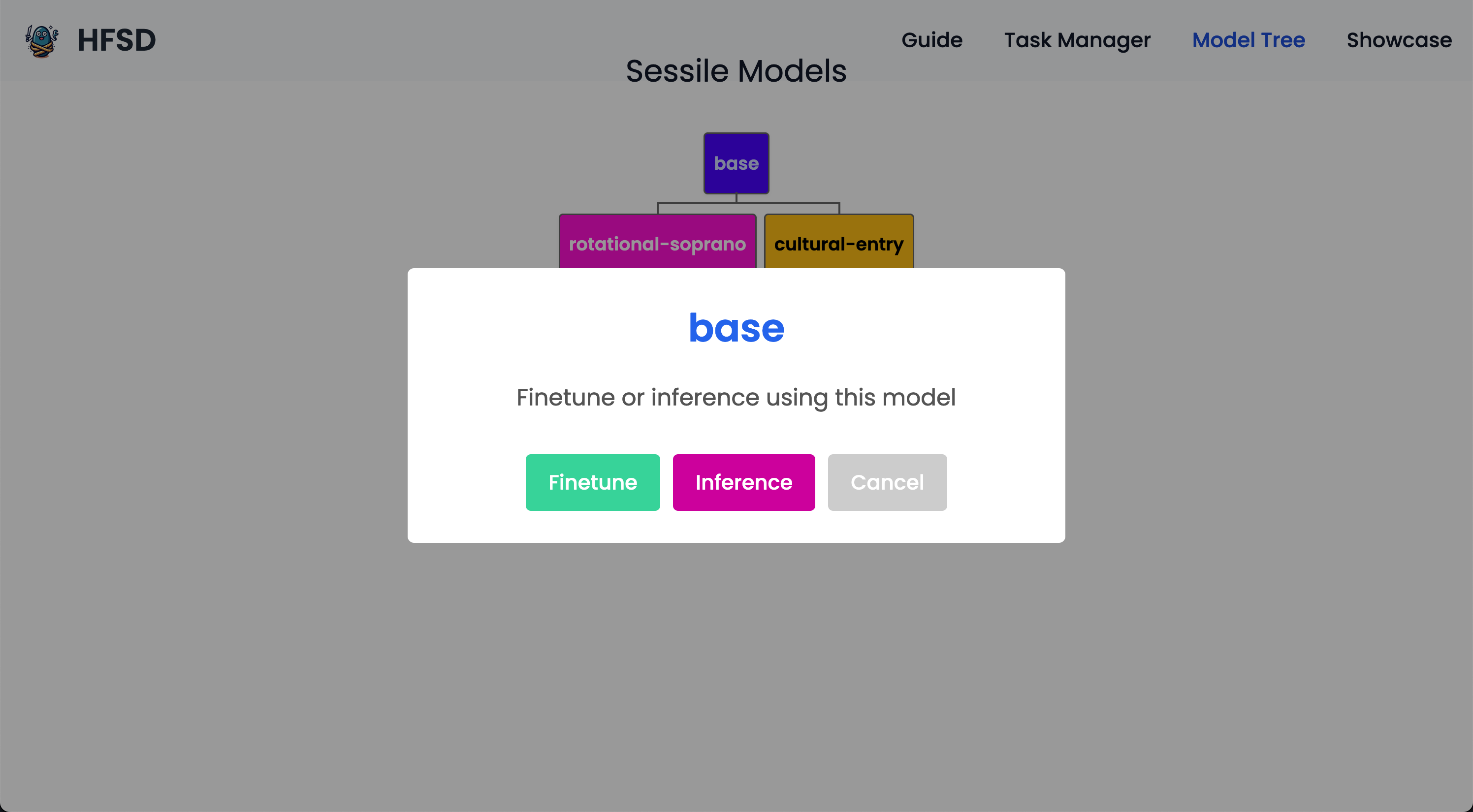}
    \end{minipage}
    \vspace{-5mm}
    \caption{%
        Overview of Model Tree and options when clicking each trained node.
    }\label{fig:webapp-tree_option}
    \Description{A model tree.}
    \vspace{-5mm}
\end{figure}

\paragraph{Model Tree}
Since our approach in fine-tuning the model led to a significant improvement in the performance of the LoRA model over time, we can conceptualize each successive version of the model as a node within a hierarchical tree structure. This tree structure effectively visualizes the evolutionary progression of our model development. As depicted in Fig.~\ref{fig:webapp-tree_option}, the tree is organized in a top-to-bottom fashion. The root node, situated at the apex of the tree, represents the initial baseline model prior to any fine-tuning. Each subsequent node descending from the root represents a new iteration of the model, refined and enhanced through successive stages of fine-tuning. The hierarchical nature of this tree allows for a clear and systematic representation of the iterative improvements made to the model, highlighting the progressive enhancement in its performance and capabilities. Each node in the tree thus signifies a specific version of the model, encapsulating the modifications and optimizations that have been applied up to that point. This visualization not only aids in understanding the developmental trajectory of our model but also underscores the cumulative impact of our fine-tuning approach in systematically advancing the model's efficacy.

\begin{figure}[t!]
    \begin{minipage}[t]{\linewidth}
        \centering
        \includegraphics[width=\linewidth]{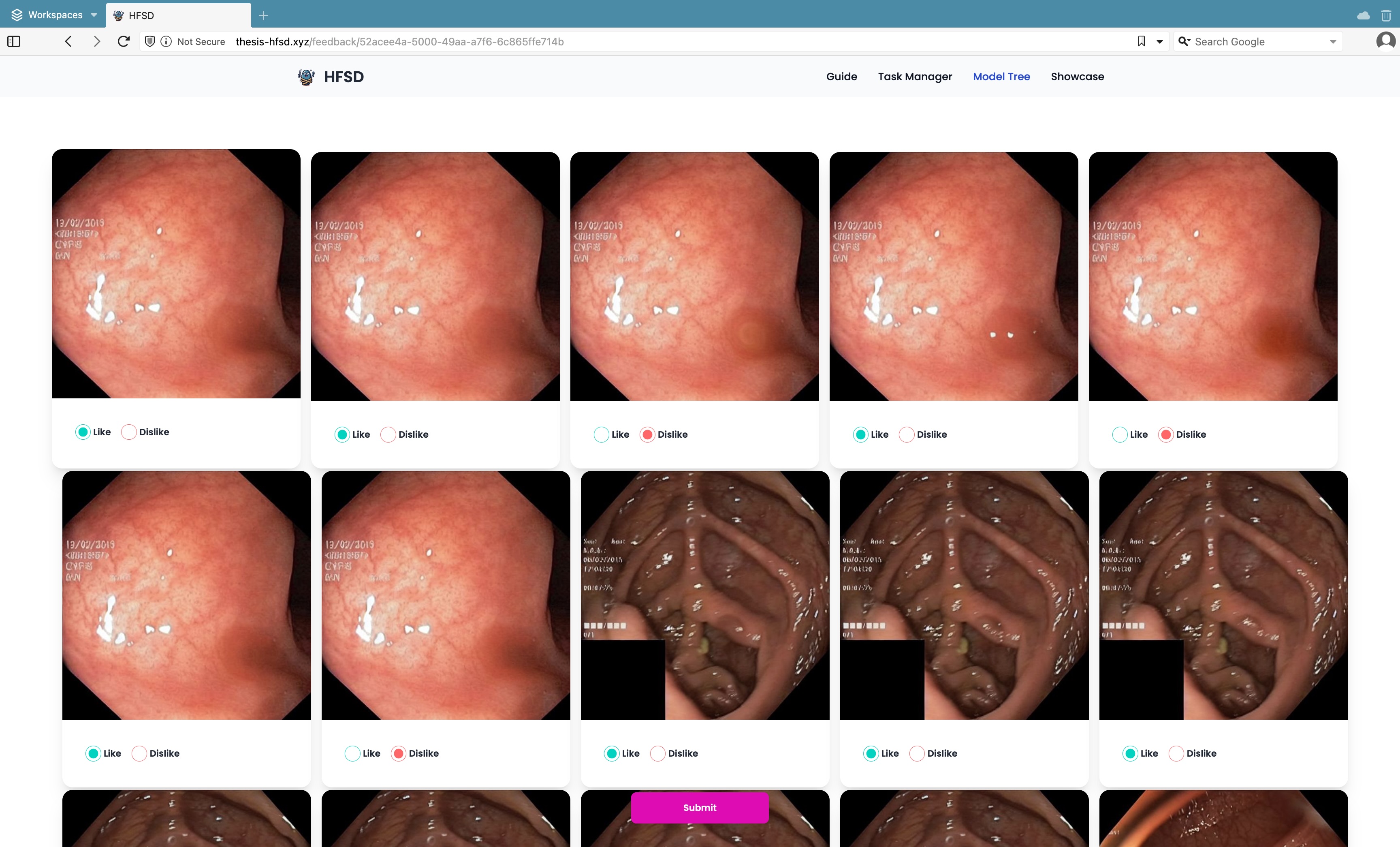}
    \end{minipage}

    

    
    
    
    \vspace{-3mm}
    \caption{%
        Sampling of polyp image dataset.
    }\label{fig:webapp-sample}
    \Description{A page sample 2.}
\end{figure}

\begin{figure}[t!]
    \centering
    \includegraphics[width=\linewidth]{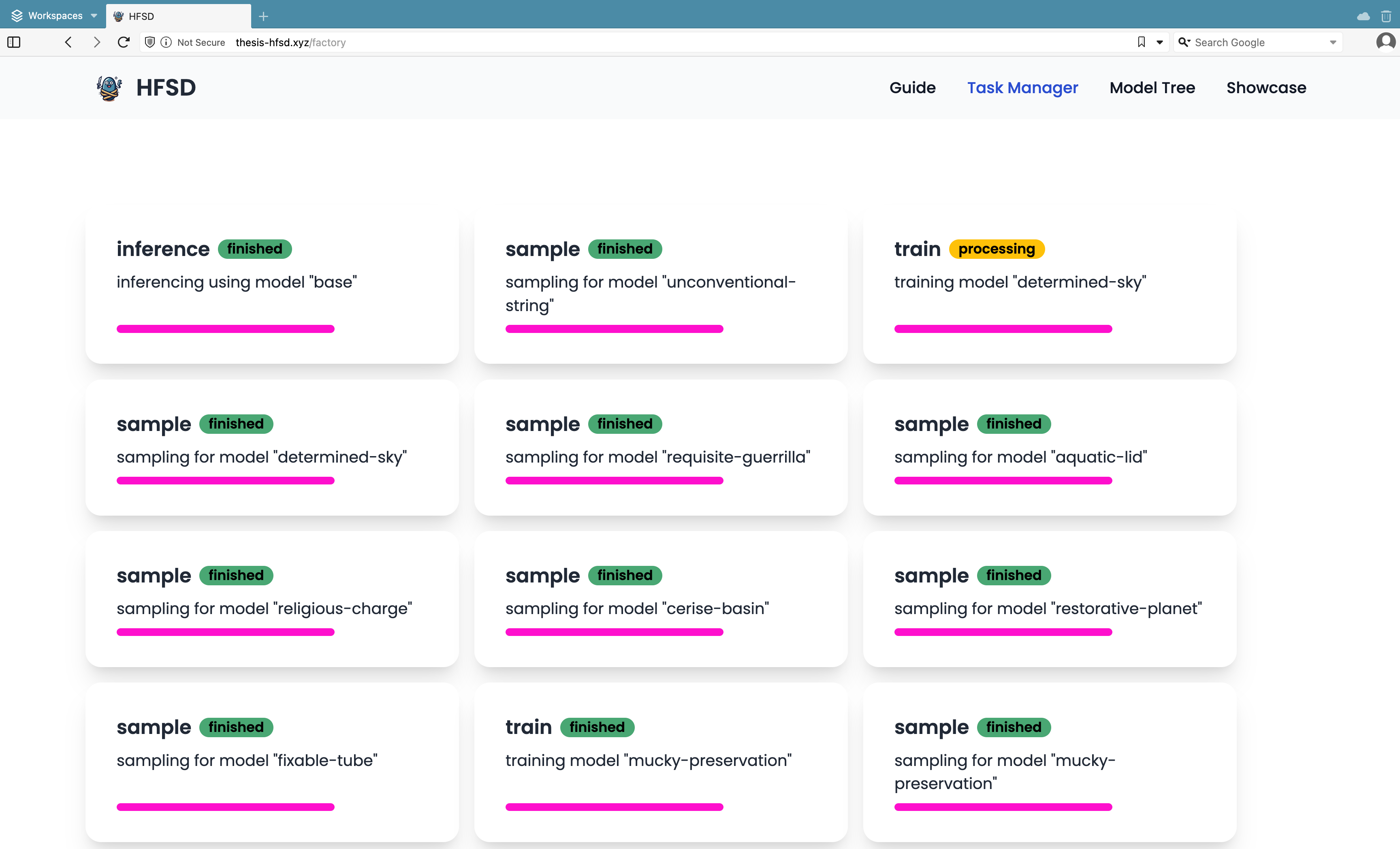}
    \vspace{-3mm}
    \caption{Task Manager for tracking current tasks.}
    \label{fig:task-manager}
    \Description{A task manager.}
    \vspace{-5mm}
\end{figure}

\paragraph{Sampling and Feedback Collection}
Upon clicking on any trained model node (as illustrated in Fig.~\ref{fig:webapp-tree_option}), two additional actions become available. Firstly, we can choose to continue fine-tuning the selected model, effectively creating a new iteration that branches off from the current node. This action allows us to further refine and optimize the model, thereby generating a new version that reflects the ongoing improvements. Secondly, we have the option to use the selected model for inference with our dataset. This functionality enables us to apply the trained model to real-world data, evaluating its performance and gaining insights based on its predictions. Both of these actions, continuing the fine-tuning process and utilizing the model for inference, provide crucial flexibility in our workflow, facilitating continuous enhancement and practical application of our models.

Regarding the sampling process, as depicted in Fig.~\ref{fig:webapp-sample}, the user interface is intentionally designed to be simple and straightforward, offering users the ability to rate the generated images with either a "like" or "dislike." This binary feedback system aligns with our methodology, which leverages user preferences to enhance the model. The feedback is mapped to values of 0 for "like" and -1 for "dislike", streamlining the user experience and minimizing distractions from a potentially complex system. This approach ensures that users can provide robust and meaningful feedback without being overwhelmed by excessive options.

Once the user has rated each image in the sample set, the feedback is collected and ready to be used in the model training process. Users can initiate this process by clicking the "Submit" button located at the bottom of the page. This seamless flow from rating to submission helps maintain user engagement and supports an efficient and user-friendly feedback loop, ultimately contributing to the iterative improvement of our model.

\begin{figure}[t!]
    \begin{minipage}[t]{\linewidth}
        \centering
        \includegraphics[width=\linewidth]{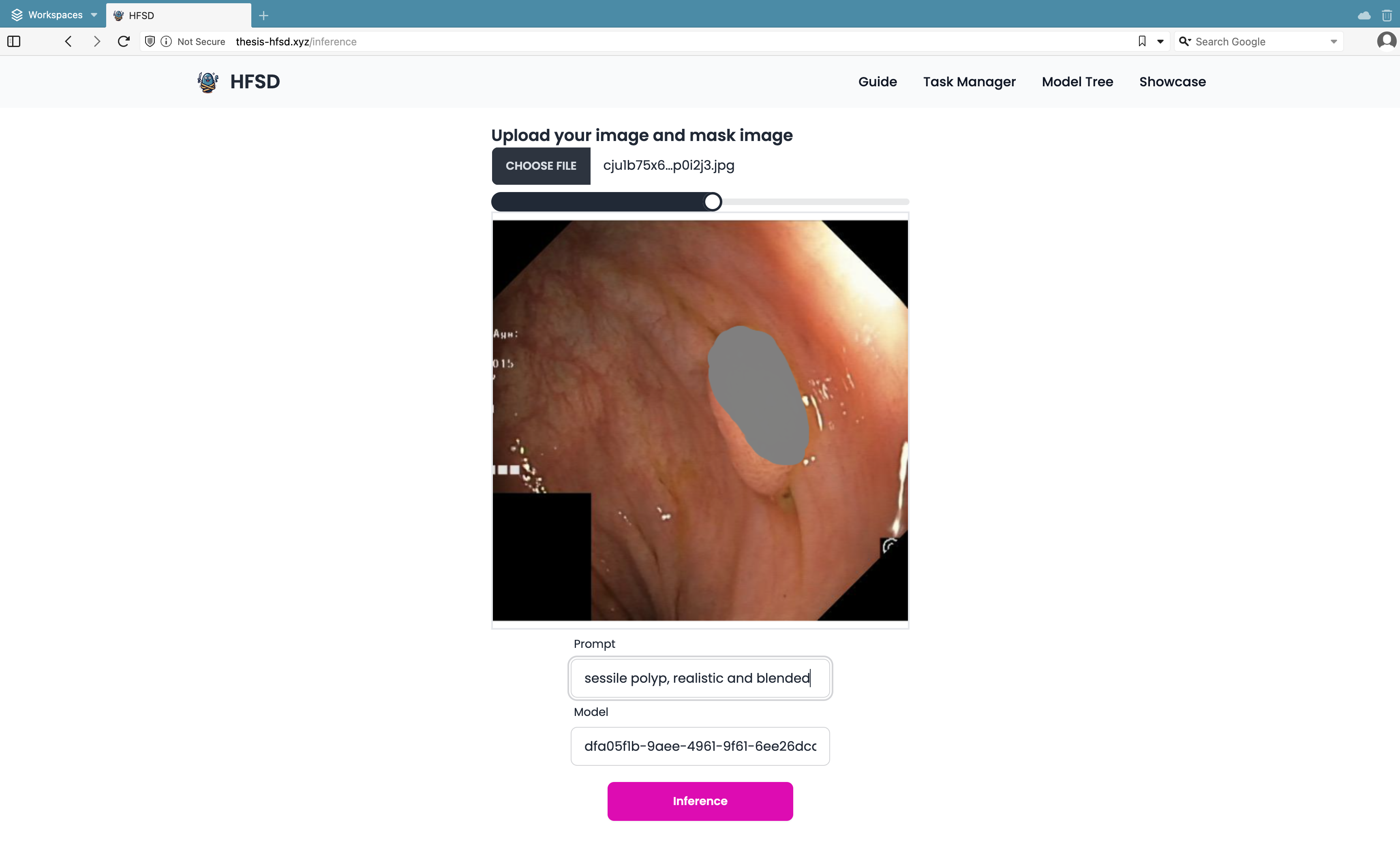}
    \end{minipage}
    \hspace{0.1cm}
    \begin{minipage}[t]{\linewidth}
        \centering
        \includegraphics[width=\linewidth]{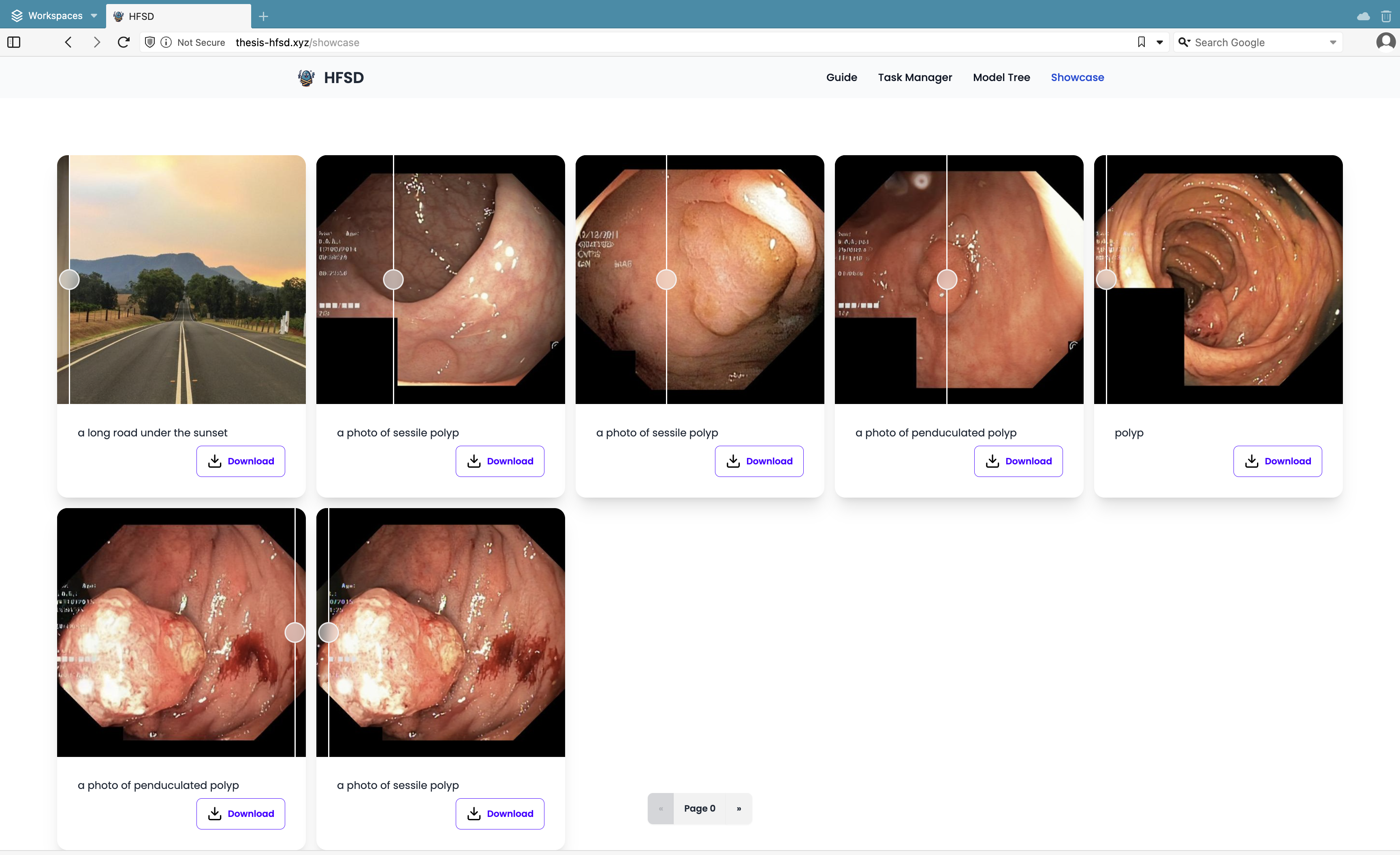}
    \end{minipage}
    \vspace{-3mm}
    \caption{%
        Inference step and showcase interface.
    }\label{fig:webapp-inference_showcase}
    \Description{A showcase interface.}
    \vspace{-5mm}
\end{figure}

\subsubsection{Inference with model}
After selecting the desired model for inference, the user is directed to the screen of inference step. Here, the user first selects an image from their file system. To support image inpainting, the user must draw a mask on the image to specify the region for modification. A default brush is available for drawing the mask, with a slider to adjust the brush size for precision. Next, the user provides a text prompt describing the desired modifications.

The final screen of our service is dedicated to showcasing the inference outputs for every user request. This screen displays both the input and output images along with the prompts provided by each user to the model. By doing so, it allows users to see the transformations and modifications performed by the model in response to specific prompts.

The inference step and showcase interface can be seen in Fig.~\ref{fig:webapp-inference_showcase}.

\subsubsection{Task Manager}
After a user submits a request for a sample, fine-tuning, or inference from our service, each task requires a variable amount of time to complete, depending on the complexity and nature of the request. Inference tasks typically finish relatively quickly, taking approximately 10 seconds, whereas model training tasks are more time-consuming, often requiring up to 20 minutes to complete.

To facilitate transparency and keep users informed about the status of their requests, the website includes a task manager feature, as illustrated in Fig.~\ref{fig:task-manager}. This task manager provides real-time updates on the progress of each request, allowing users to monitor the status and estimated completion time. The interface is designed to be user-friendly, displaying a list of active tasks along with their current stages of processing. This ensures that users are not left in the dark during longer operations and can plan their time accordingly.

\section{Experiments}

\begin{figure*}[t!]
    \centering
    
    \begin{subfigure}[b]{0.49\linewidth}
        \centering
        \includegraphics[width=0.8\linewidth]{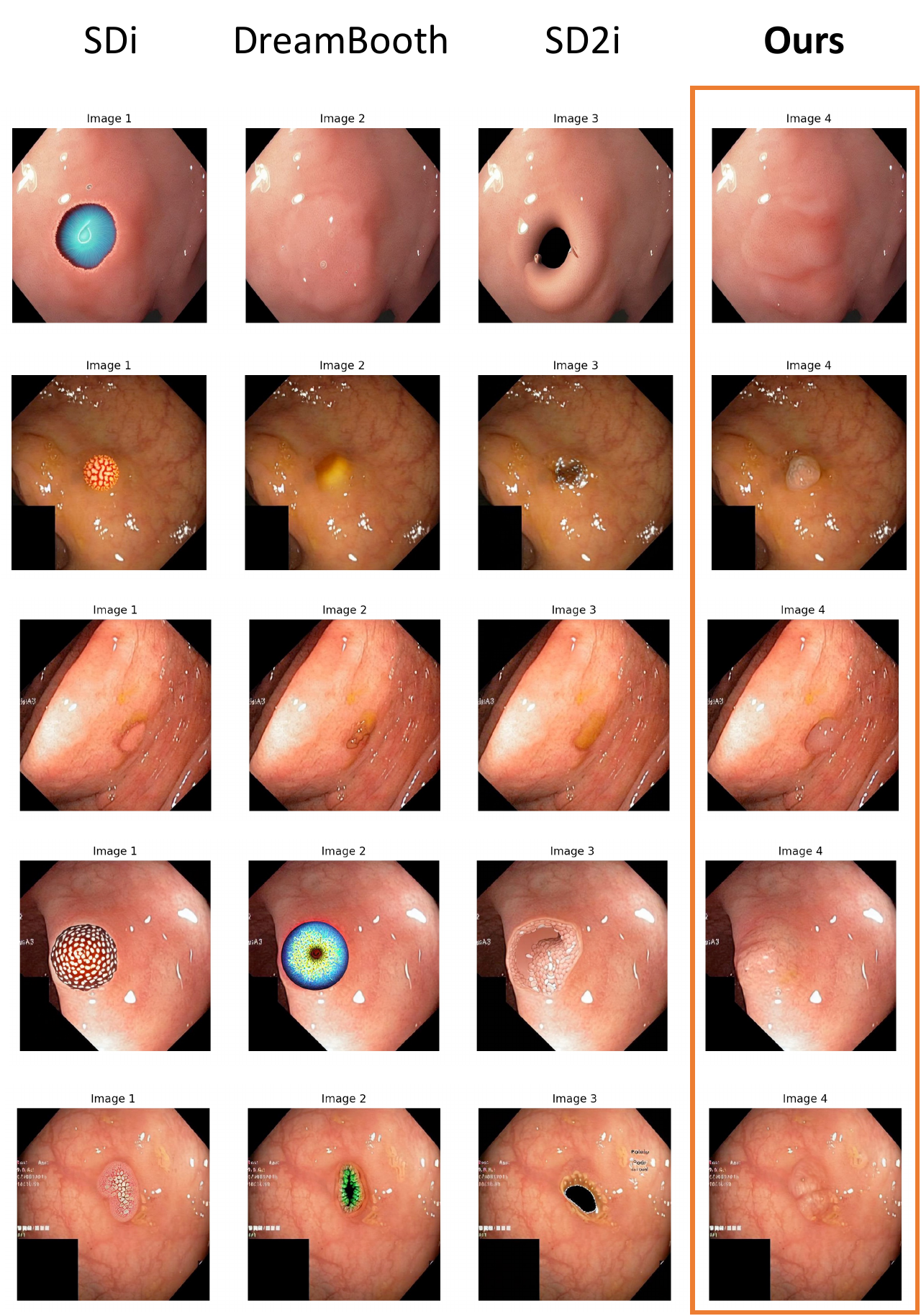}
        \caption{Sessile Polyps.}
        \label{fig:exp-qualitative-sessile}
    \end{subfigure}
    \hfill
    \begin{subfigure}[b]{0.49\linewidth}
        \centering
        \includegraphics[width=0.8\linewidth]{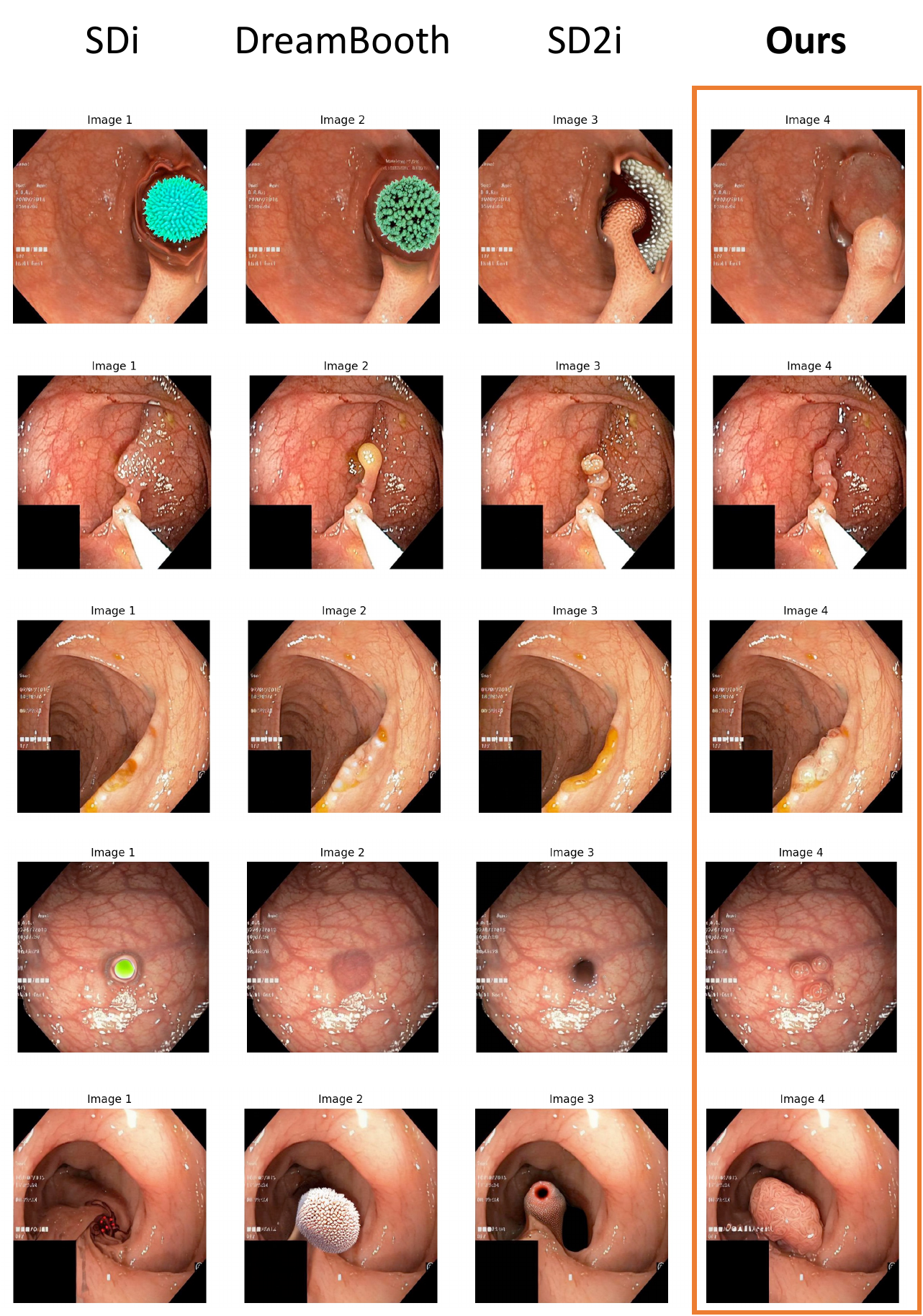}
        \caption{Pedunculated Polyp.}
        \label{fig:exp-qualitative-pedunculated}
    \end{subfigure}
    
    \caption{Qualitative comparison of inpainting results across different polyp subsets.}
    \label{fig:exp-qualitative-combined}
    \Description{A side-by-side qualitative comparison of output images showing inpainting results for both sessile and pedunculated polyps.}
\end{figure*}

\begin{table}[t!]
    \centering
    \caption{Evaluation results on the Sessile Polyp subset.}
    \label{tab:exp-quantitative-sessile}
    \resizebox{\linewidth}{!}{%
        \begin{tabular}{l|ccccc}
            \toprule
            \textbf{Model} & \textbf{L1 Error} $\downarrow$ & \textbf{L2 Error} $\downarrow$ & \textbf{SSIM} $\uparrow$ & \textbf{PSNR} $\uparrow$ & \textbf{LPIPS} $\downarrow$ \\
            \midrule
            SDi (base) & 0.0267 & 0.0044 & 0.8761 & 36.1593 & 0.8202 \\
            DreamBooth & 0.0266 & 0.0041 & 0.8776 & 36.3210 & 0.8233 \\
            SD2i & 0.0227 & 0.0037 & 0.8843 & 37.2751 & 0.8586 \\
            \textbf{Ours} & \textbf{0.0169} & \textbf{0.0033} & \textbf{0.8883} & \textbf{37.2926} & \textbf{0.8176} \\
            \bottomrule
        \end{tabular}
    }
    
\end{table}

\subsection{Dataset}

To evaluate our method, we utilized Kvasir-SEG~\cite{jha2020kvasir}, an open-access dataset of gastrointestinal polyp images featuring ground-truth segmentation masks manually annotated and verified by an experienced gastroenterologist. Built upon the Kvasir Dataset v2, the images vary in resolution from 332$\times$487 to 1920$\times$1072 pixels. Specifically, our experimental setup incorporated 3,842 sessile polyp images and 1,034 pedunculated polyp images.

\subsection{Evaluation Metrics}

Image Quality Assessment (IQA) quantifies the visual degradation of a generated image by comparing it against an ideal reference, capturing both objective technical deviations and subjective human perception \cite{thung2009survey}. To rigorously evaluate our inpainting results, we employ a comprehensive suite of standard metrics: L1 error, L2 error, Structural Similarity Index (SSIM), Peak Signal-to-Noise Ratio (PSNR), and Learned Perceptual Image Patch Similarity (LPIPS).

\subsection{Compared Methods and Setup}

To thoroughly evaluate our approach, we compared our method against leading diffusion-based image inpainting techniques. The methods included in this assessment were the fine-tuning technique DreamBooth \cite{ruiz2023dreamboothfinetuningtexttoimage}, Stable Diffusion Inpainting (SDi) \cite{Rombach_2022_CVPR}, and its successor, Stable Diffusion 2 Inpainting (SD2i). For our experiments, we actively fine-tuned both our proposed method and DreamBooth, while evaluating the two baseline Stable Diffusion models (SDi and SD2i) in their unmodified, pre-trained states.

\subsection{Results}

\begin{table}[t!]
    \centering
    \caption{Evaluation results on the Pedunculated Polyp subset.}
    \label{tab:exp-quantitative-pedunculated}
    \resizebox{\linewidth}{!}{%
        \begin{tabular}{l|ccccc}
            \toprule
            \textbf{Model} & \textbf{L1 Error} $\downarrow$ & \textbf{L2 Error} $\downarrow$ & \textbf{SSIM} $\uparrow$ & \textbf{PSNR} $\uparrow$ & \textbf{LPIPS} $\downarrow$ \\
            \midrule
            SDi (base) & 0.0458 & 0.0118 & 0.8063 & 31.3990 & 0.7096 \\
            DreamBooth & 0.0456 & 0.0115 & 0.8074 & 31.3647 & 0.7116 \\
            SD2i & 0.0434 & 0.0114 & 0.8117 & 31.4341 & 0.7316 \\
            \textbf{Ours} & \textbf{0.0359} & \textbf{0.0110} & \textbf{0.8197} & \textbf{31.5410} & \textbf{0.6675} \\
            \bottomrule
        \end{tabular}
    }
    
\end{table}


When evaluated on the \textit{Sessile Polyp} subset, our method shows marked improvements in L1 and L2 errors alongside elevated SSIM scores, demonstrating highly accurate structural and pixel-level synthesis (Table~\ref{tab:exp-quantitative-sessile}). The accompanying reduction in LPIPS scores further confirms enhanced perceptual realism, a crucial factor for medical imagery. Similarly, on the \textit{Pedunculated Polyp} subset, our approach outperforms all baselines by achieving the best scores across every metric (Table~\ref{tab:exp-quantitative-pedunculated}). Attaining the lowest L1 and L2 errors, the highest SSIM, and the lowest LPIPS scores confirms that our expert-guided method delivers comprehensive advancements in both visual fidelity and quantitative precision.

In addition to our quantitative evaluation, we conducted a comprehensive qualitative assessment of the generated inpainting results across both the pedunculated and sessile polyp images. Visual comparisons against baseline techniques reveal that our proposed method successfully minimizes the structural inconsistencies and blending artifacts commonly produced by standard diffusion models. As demonstrated in Fig.~\ref{fig:exp-qualitative-combined}, while baseline methods often struggle with unnatural textures or poor boundary integration, our expert-in-the-loop approach consistently generates highly realistic, anatomically appropriate polyp images with seamless transitions. This qualitative improvement directly highlights the value of leveraging expert medical feedback to guide the generative process, ensuring the outputs are contextually accurate for clinical applications.

\section{Pilot Study}

\begin{figure}[t!]
    \centering
    
    \begin{subfigure}[b]{0.32\linewidth}
        \centering
        \includegraphics[width=\linewidth]{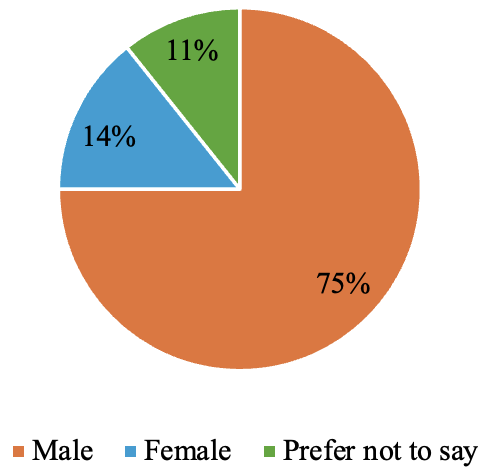}
        \caption{Gender}
        \label{fig:userstudy-gender}
    \end{subfigure}
    \hfill
    \begin{subfigure}[b]{0.32\linewidth}
        \centering
        \includegraphics[width=\linewidth]{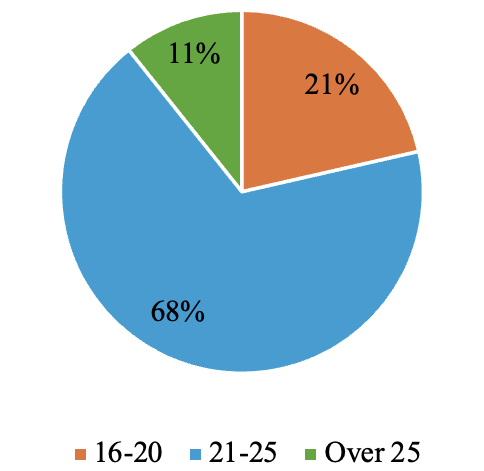}
        \caption{Age}
        \label{fig:userstudy-age}
    \end{subfigure}
    \hfill
    \begin{subfigure}[b]{0.32\linewidth}
        \centering
        \includegraphics[width=\linewidth]{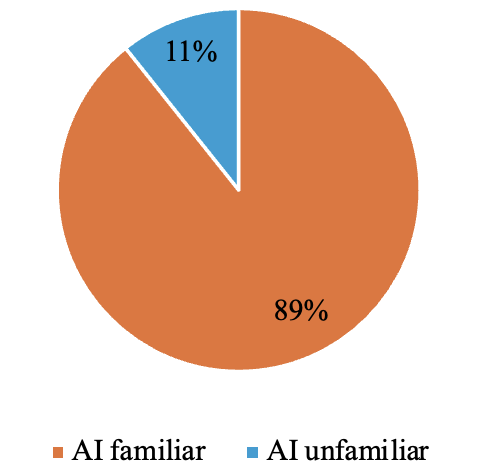}
        \caption{AI background}
        \label{fig:userstudy-background}
    \end{subfigure}
    
    \caption{Information of participants.}
    \label{fig:userstudy-participants}
    \Description{Three pie charts displaying the demographic breakdown of the 28 study participants by Gender, Age, and AI background.}
\end{figure}

\begin{figure*}[t!]
    \centering
    \includegraphics[width=\textwidth]{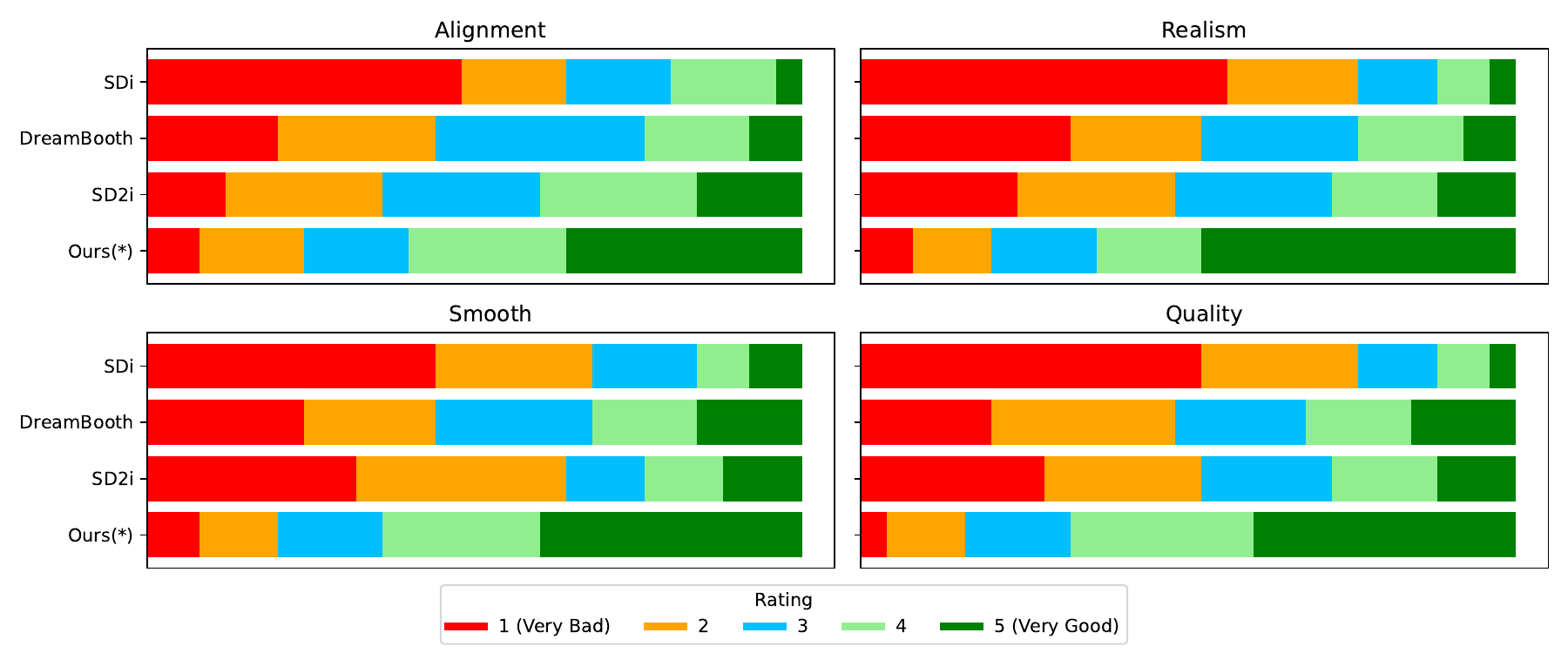}
    \caption{Statistical distribution of user ratings for each image inpainting method. Our proposed method consistently outperforms all other methods across all evaluated metrics.}
    \label{fig:userstudy-overall}
    \Description{A user study chart.}
    \vspace{-3mm}
\end{figure*}

\subsection{Participants}

We invited 28 participants (21 males, 4 females, 3 unknown; aged between 16 and 30) from our research community for our study. Almost participants are familiar with AI, including 3 medical doctors, 20+ computer science undergraduate students, and those from other fields. Figure~\ref{fig:userstudy-participants} illustrates details of the participants. With their diverse professional backgrounds, they contributed different viewpoints to the evaluation process, ensuring a well-rounded and objective assessment. 

\subsection{Dataset}

We randomly selected 20 images from each category of sessile and pedunculated polyp images.

\subsection{Evaluation Metrics}

To ensure a comprehensive evaluation of methods, we established four key metrics: alignment, realism, smoothness, and quality. These metrics allowed us to thoroughly assess the effectiveness and capabilities of each method.
\begin{itemize}
    \item \textit{Alignment}: Assessing how well the edited image matches the original intent and structure described by the user prompt.
    \item \textit{Realism}: Evaluating the natural appearance and believability of the edits, ensuring that the modifications blend seamlessly with the original image.
    \item \textit{Smoothness}: Analyzing the smoothness and continuity of the transitions between edited and non-edited parts of the image.
    \item \textit{Quality}: Gauging the overall visual quality of the edited image, including clarity, coherence, and aesthetic appeal.
\end{itemize}

\subsection{Apparatus and Procedure}


Participants rated the performance of each of the four methods on a scale of 1 ("very bad") to 5 ("very good") across four metrics based on their perspectives. The evaluation was carried out using a custom-built web-based interface that displayed images in real time. 

The pilot study was conducted both online and in-person to ensure diverse participation. Online participants were given detailed instructions and a tutorial session to familiarize them with the evaluation process, while in-person participants received direct assistance if needed. Data collected was securely stored and later analyzed to produce the Mean Opinion Scores (MOS) for each metric and method combination.

\subsection{Quantitative Results}

\begin{table}[t!]
    \centering
    \caption{Overall user study results comparing image inpainting methods using Mean Opinion Scores (MOS). Our proposed method comprehensively outperforms all baseline methods across all tested categories.}
\resizebox{\linewidth}{!}{
    \begin{tabular}{l|cc|cc}
        \toprule
        \multirow{2}{*}{\textbf{Method}} 
        & \multicolumn{2}{c|}{\textbf{Non-Medical Expert}} 
        & \multicolumn{2}{c}{\textbf{Medical Expert}} \\
        \cmidrule{2-5}
        & \shortstack{\textbf{Sessile}\\\textbf{Polyp}} 
        & \shortstack{\textbf{Pedunculated}\\\textbf{Polyp}} 
        & \shortstack{\textbf{Sessile}\\\textbf{Polyp}} 
        & \shortstack{\textbf{Pedunculated}\\\textbf{Polyp}} \\
        \midrule
        SDi (base)         & 2.808 & 2.960  & 1.636 & 1.636 \\
        DreamBooth  & 3.240 & 3.176  & 1.818 & 1.955 \\
        SD2i        & 3.328 & 3.248  & 1.818 & 2.045 \\
        \textbf{Ours}& \textbf{4.376} & \textbf{4.344}  & \textbf{3.727} & \textbf{3.545} \\
        \bottomrule
    \end{tabular}
    }
    
    \label{tab:userstudy-results}
\end{table}

\begin{figure}[t!]
    \centering
    \includegraphics[width=\linewidth]{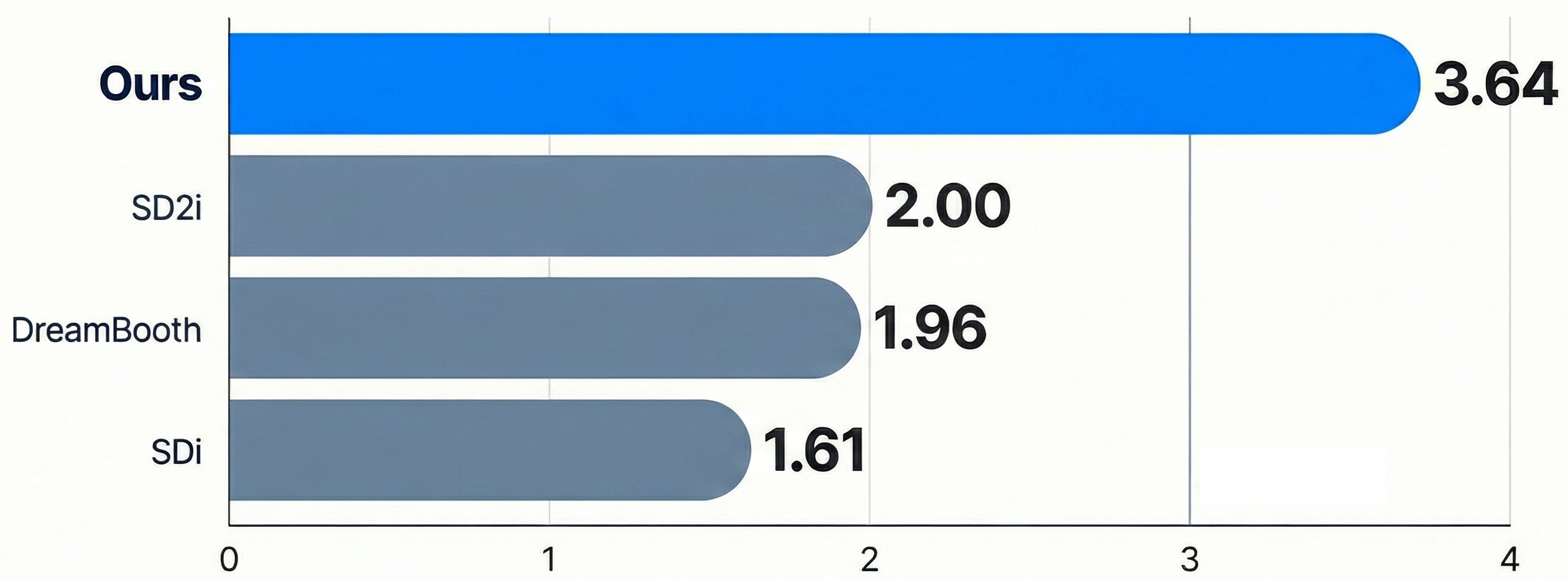}
    \caption{Overall expert clinical validation ratings assigned by specialized medical doctors. Our proposed method demonstrates superior expert-guided accuracy, outperforming existing baselines by nearly double.}
    \label{fig:userstudy-mos-doc}
    \vspace{-3mm}
\end{figure}

\subsubsection{Overall Participant Evaluation}

To evaluate the general performance of our method, we analyzed the average rating scores provided by all participants. As shown in Table~\ref{tab:userstudy-results} and Fig.~\ref{fig:userstudy-overall}, our proposed approach consistently outperforms all baseline methods across every category. We also obtained insightful conclusions from the study:
\begin{itemize}
    \item \textbf{Alignment}: Our method overwhelmingly received the highest rating score of 5, indicating a high level of alignment with user prompts. In contrast, SD2i and DreamBooth yielded dispersed scores, and SDi predominantly received scores of 1.
    
    \item \textbf{Realism}: Participants found our generated outputs to be highly realistic, predominantly awarding our method a score of 5, while baseline methods exhibited more variance with a larger proportion of low ratings.
    
    \item \textbf{Smoothness}: Our approach led this category with the majority of ratings being 5. DreamBooth and SD2i produced mixed results, and SDi frequently scored 1, indicating noticeable smoothness issues.

    \item \textbf{Quality}: Our method maintained its lead by predominantly scoring 5, whereas SDi performed the poorest, accumulating a significant number of 1s and 2s.
    
\end{itemize}

\subsubsection{Medical Expert Validation} 

Beyond general user feedback, we specifically analyzed evaluations from specialized medical doctors to assess clinical viability. As presented in Fig.~\ref{fig:userstudy-mos-doc}, this expert validation clearly indicates that our approach significantly outperforms all other evaluated methods. Notably, medical experts applied substantially stricter grading criteria compared to non-experts, yet our method still demonstrated superior performance across both polyp datasets compared to existing baselines (see Table~\ref{tab:userstudy-results}). By leveraging direct feedback from medical professionals, our model successfully avoids generating the inaccurate anatomical artifacts commonly produced by standard generative models. This expert-guided validation ensures the production of highly realistic, reliable synthetic polyp images that are vital for accurate disease diagnosis and treatment planning. Furthermore, by utilizing D3PO to bypass computationally expensive reward models, this fine-tuning process becomes highly accessible and practical for real-world medical institutions.

\begin{figure}[t!]
    \centering
    \includegraphics[width=\linewidth]{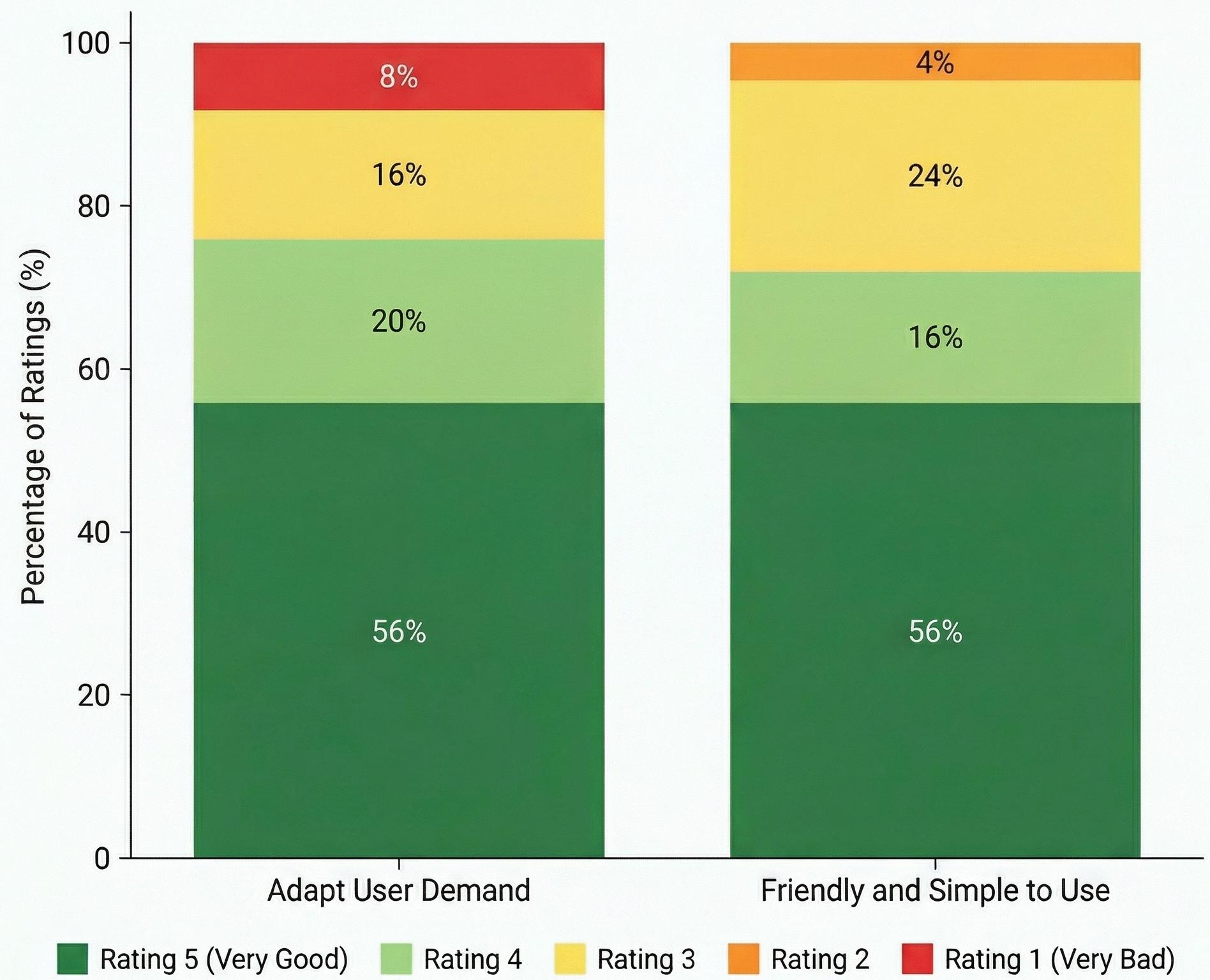}
    \caption{User survey feedback regarding the usability and adaptability of the web-based interface.}
    \label{fig:webapp-survey}
    \Description{Survey for web app.}
    \vspace{-3mm}
\end{figure}

\subsubsection{Web-Based Interface Evaluation}

To gauge user satisfaction and gather feedback on the effectiveness of our interactive platform, we conducted a comprehensive survey. Participants evaluated the system across two primary criteria: adaptability to user demands and the provision of a friendly, simple interface. Using a 5-point scale, where 1 indicates strong disagreement and 5 indicates strong agreement, users provided highly positive feedback. As visualized in Fig.~\ref{fig:webapp-survey}, the responses indicate a high level of participant satisfaction with our proposed system across both usability and adaptability dimensions.

\section{Conclusion}

This paper explores the integration of human feedback into image generation models, focusing on medical image inpainting. By incorporating expert feedback into the training process, we proposed a system that improves the realism of generated images compared to traditional models. This platform allows experts, even those with limited IT skills, to provide valuable insights into the inpainting process, ensuring that the models are refined based on real-world expertise and preferences. Extensive experiments indicated that our method significantly outperforming existing methods, showing improvements in visual coherence and contextual integration of our proposed method.

Future work involves several promising avenues to further advance image inpainting within human feedback supporting, such as improving the prompt generation to better suit the specific domains being sampled, extending the functionality of our web-based interface to support a broader range of feedback types and more interactive features.



\section*{Acknowledgments}

This research is funded by Vietnam National University - Ho Chi Minh City (VNU-HCM) under Grant Number B2026-18-17. 


\bibliographystyle{ACM-Reference-Format}
\balance
\bibliography{sample-base}

\end{document}